\documentclass[10pt,conference]{IEEEtran}

\usepackage[normalem]{ulem}

\usepackage{cite}
\usepackage{amssymb}

\ifCLASSINFOpdf
   \usepackage[pdftex]{graphicx}
   \graphicspath{{figs/}}
   \DeclareGraphicsExtensions{.pdf,.jpeg,.png}
\else
   \usepackage[dvips]{graphicx}
   \graphicspath{{../figs/}}
   \DeclareGraphicsExtensions{.eps}
\fi
\usepackage[cmex10]{amsmath}
\usepackage{amsthm}

\usepackage{algorithmic}

\usepackage{array}

\usepackage[english]{babel}
\usepackage{hyperref}

\addto\extrasenglish{%
}

\ifCLASSOPTIONcompsoc
  \usepackage[caption=false,font=normalsize,labelfont=sf,textfont=sf]{subfig}
\else
  \usepackage[caption=false,font=footnotesize]{subfig}
\fi
\usepackage{url}
\usepackage{tikz}
\usetikzlibrary{shapes.geometric, arrows.meta, positioning}

\usepackage{xspace}

\newif\iffinal

\finaltrue

\newcommand{\cmtid}{241}

\newcommand{\myparagraph}[1]{\smallskip\noindent\textbf{#1}\xspace}

\iffinal
\else
\usepackage[switch]{lineno}
\fi

\newcommand{\figPipelines}{%
\begin{figure*}[!t]
  \includegraphics[width=\linewidth]{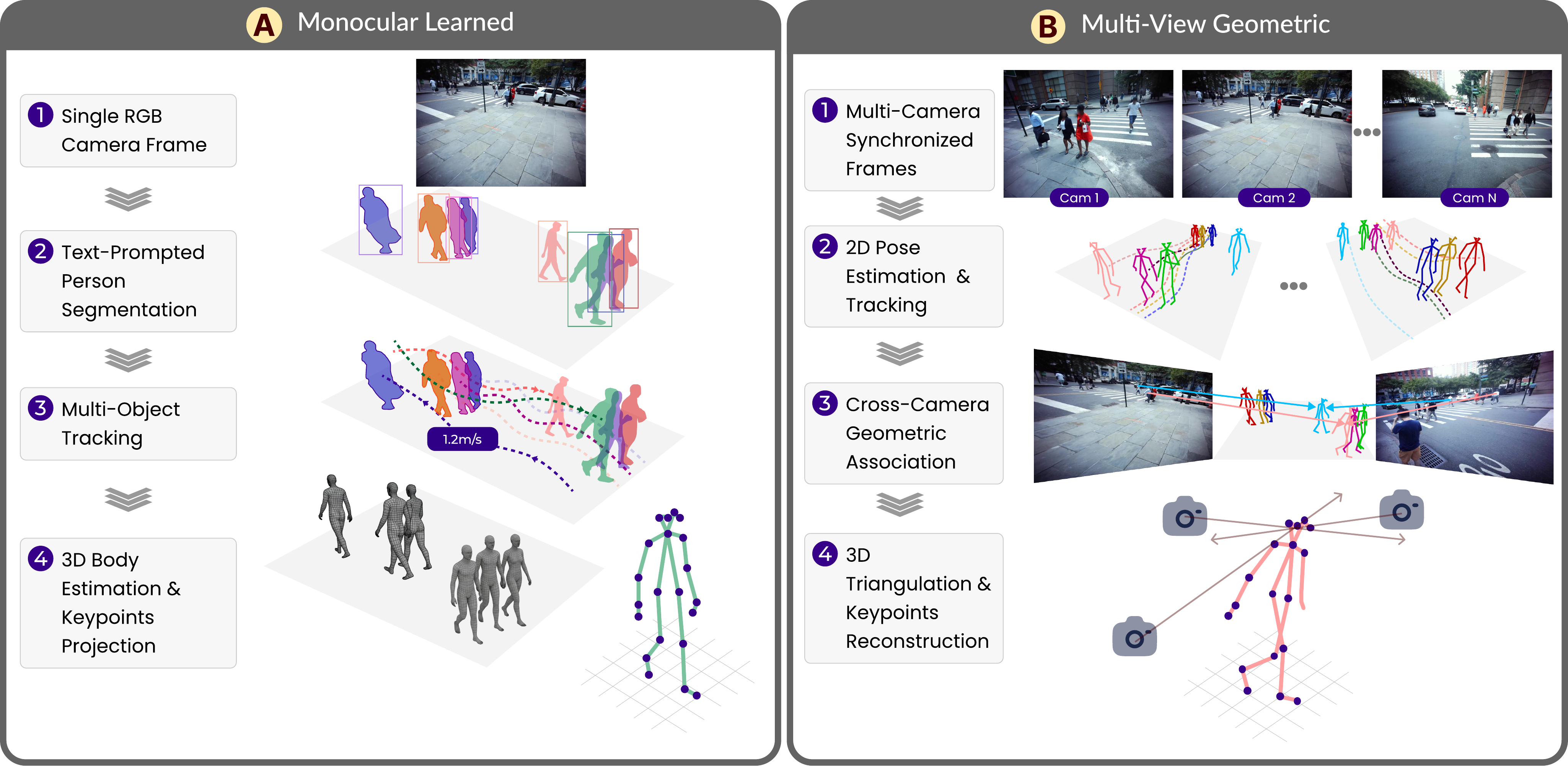}
  \caption{Overview of the two compared strategies for deriving pedestrian motion descriptors: Pipeline~A (monocular learned, left) and Pipeline~B (multi-view geometric, right).}
  \label{fig:pipelines}
  \vspace{-0.4cm}
\end{figure*}}

\newcommand{\figMorphology}{%
\begin{figure*}[t]
  \centering
  \includegraphics[width=\textwidth]{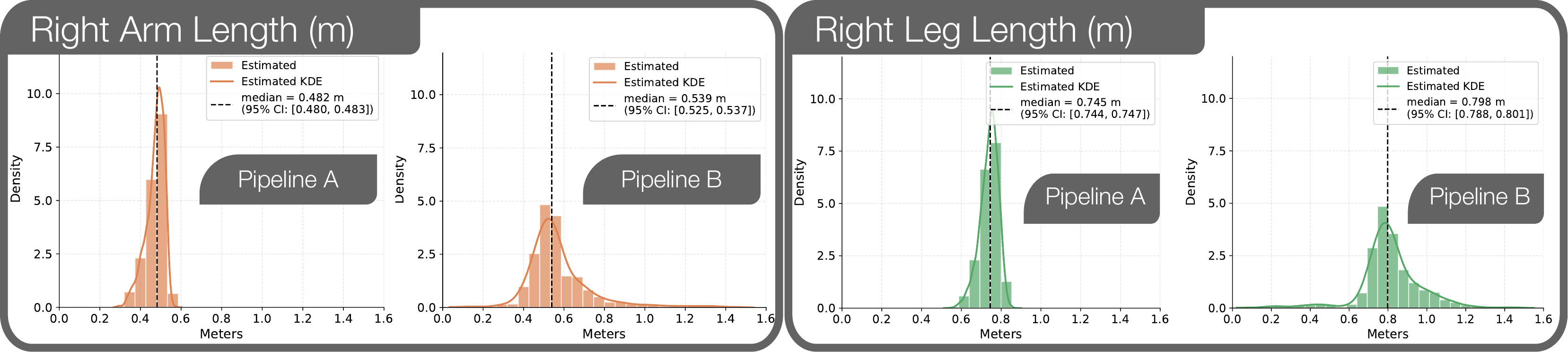}
  \caption{Comparison of reconstructed morphological metrics (right leg length, top; right arm length, bottom) between Pipeline~A (left) and Pipeline~B (right). The distributions are estimated using a per-tracklet global median assignment, with dashed lines indicating the median values.}
  \label{fig:morfologia}
\end{figure*}}

\newcommand{\figSpeedAnalysis}{%
\begin{figure}[!t]
  \centering
  \includegraphics[width=\linewidth]{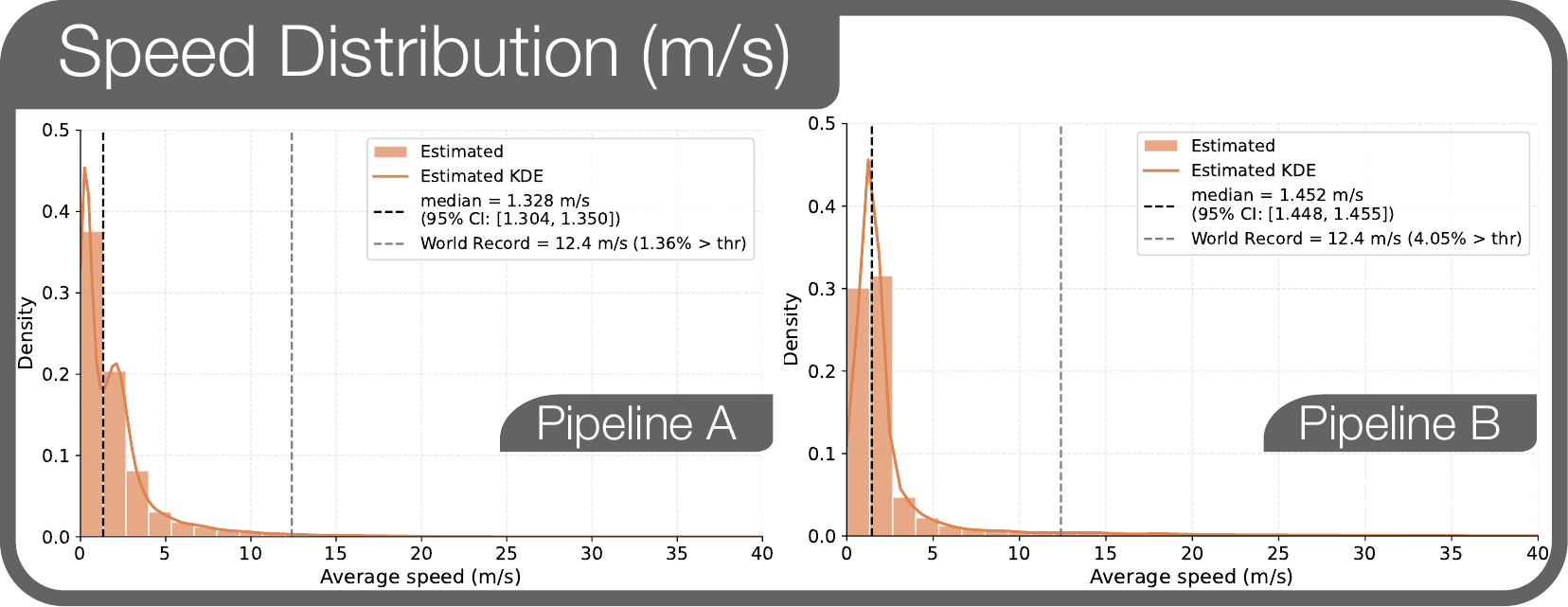}
  \caption{Comparison of estimated pedestrian speed distributions between Pipeline~A (left) and Pipeline~B (right). Histograms show the estimated density with overlaid kernel density estimates (KDE); dashed lines indicate the median speed values (1.328~m/s for Pipeline~A and 1.452~m/s for Pipeline~B).}
  \label{fig:velocity_comparison}
\end{figure}}

\newcommand{\figStatureComparison}{%
\begin{figure}[!t]
  \centering
  \includegraphics[width=\linewidth]{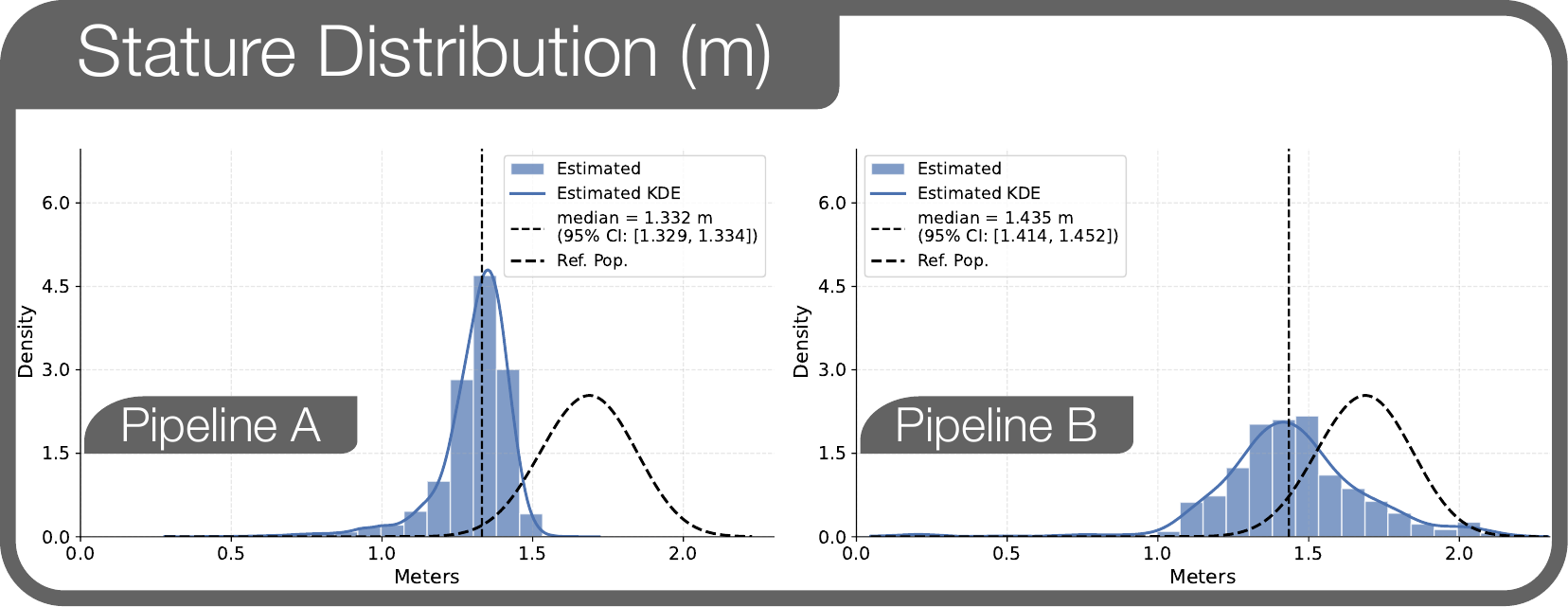}
  \caption{Comparison of estimated total stature distributions across all three sessions between Pipeline~A (monocular learned, left) and Pipeline~B (multi-view geometric, right). The dashed curve representing the reference population (\textit{Ref. Pop.}) models the adult population stature as an equal mixture of male and female normals parameterized from CDC data. Dashed vertical lines indicate the median of the estimated distributions.}
  \label{fig:stature_comparison}
\end{figure}}

\begin{document}
%
\title{Comparing Monocular and Multi-View 3D Pedestrian Reconstruction Strategies for Urban Intersection Sensing}


\iffinal

\author{
\IEEEauthorblockN{
Kauan Mariani$^{*}$,
Matheus Carvalho$^{*}$,
Joel Perca, João Rulff and Jorge Poco
}
\IEEEauthorblockA{
Fundação Getulio Vargas (FGV)\\
Rio de Janeiro, Brazil\\
Email: \{kauan.ferreira, matheus.carvalho.2\}@fgv.edu.br,
\{joel.quispe, joao.rulff, jorge.poco\}@fgv.br
}
}

\else
  \author{SIBGRAPI Paper ID: \cmtid \\ }
  \linenumbers
\fi

\maketitle

\iffinal
\begingroup
\renewcommand{\thefootnote}{*}
\footnotetext{Kauan Mariani and Matheus Carvalho contributed equally to this work (co-first authorship).}
\endgroup
\fi

\begin{abstract}
Understanding how pedestrians move and interact at urban intersections is central to traffic-safety assessment, mobility planning, and smart-city sensing. 
Such analyses rely on physically grounded descriptions of motion—3D body pose, metric trajectories, and pose-derived descriptors such as stature and walking speed. 
Recovering these from fixed roadside cameras is difficult: viewpoints are oblique and long-range, occlusions are frequent, and neither ground-truth 3D pose nor cross-camera identity labels are typically available. 
We compare two strategies for deriving pedestrian motion descriptors from a multi-camera intersection dataset: a monocular learned pipeline that combines text-prompted segmentation, tracking, and learned 3D body estimation, and a calibrated multi-view geometric pipeline that combines 2D pose estimation, tracking, cross-camera association, and triangulation. 
Because 3D ground truth is unavailable, we evaluate both with reference-free criteria relevant to intersection analysis: pedestrian coverage, anthropometric plausibility, walking-speed distributions, computational cost, and failure modes. 
The two strategies exhibit complementary trade-offs. The learned pipeline is simpler to deploy and covers more pedestrians, but suffers from scale ambiguity and systematic height compression; the geometric pipeline yields metric reconstructions more efficiently, but requires calibration, discards pedestrians seen by only one camera, and can produce severe outliers in dense crowds. 
We distill these findings into practical guidance for selecting a reconstruction strategy and configuring camera infrastructure according to the target analysis task.

\end{abstract}


\IEEEpeerreviewmaketitle


\section{Introduction}

Street intersections concentrate much of the risk and complexity of urban mobility into a small footprint, where pedestrians, cyclists, and vehicles negotiate the same space under tight timing constraints~\cite{kathuria2020evaluating, NABAVINIAKI2019239}.
Traditional analysis relies on static crash records and infrastructure models~\cite{tarko2012use}, but video-based sensing is shifting practice toward fine-grained behavioral analysis over time~\cite{saunier2007automated, perca2026urbanclipatlas, zhang2025language}.

Deriving interpretable physical measurements from raw 2D video requires restoring metric scale to measure parameters such as pedestrian height, walking speed, direction, and gait directly.
While recovering 3D info traditionally required multi-view reconstruction from synchronized cameras, learned monocular depth estimation now offers 3D body estimation from a single view, drastically simplifying sensing setup.
Whether monocular substitution holds for urban sensing tasks remains open. Multi-view systems demand multi-camera calibration and synchronization~\cite{streetaware}, whereas single-view methods lower deployment barriers if metric fidelity is comparable.

This paper addresses this gap by directly comparing multi-view and monocular 3D reconstruction strategies on dynamic urban intersection video.
Because 3D ground truth is unavailable, we evaluate both approaches using a distribution-based evaluation protocol to determine whether their estimates reproduce population characteristics. Our findings show that the optimal choice depends on the specific measurement required, providing actionable guidelines for instrumenting urban spaces.


Our contributions can be summarized as follows: (i) a comparison of monocular and multi-view strategies for deriving pedestrian motion descriptors from real urban intersection video; (ii) a reference-free evaluation protocol for assessing coverage, anthropometric plausibility, motion dynamics, computational cost, and failure modes when 3D ground truth is unavailable; and (iii) practical recommendations for selecting sensing and reconstruction configurations according to the target pedestrian-dynamics analysis task.


\section{Related Work}

\myparagraph{Pedestrian behavior and 3D reconstruction.}
Pedestrian corpora range from tracking benchmarks~\cite{benfold2011towncentre, wildtrack} to traffic corpora~\cite{cityflow} and crowd trackers~\cite{mot20}. However, infrastructure-mounted captures with overlapping multi-camera RGB remain rare and lack 3D ground truth~\cite{streetaware}. 
\emph{Monocular} methods regress body parameters from single images~\cite{ sam3dbody}, improving robustness to occlusion but suffering from scale ambiguity and depth jitter. 
\emph{Multi-view} methods exploit projective geometry via calibration, association, and triangulation to recover metric 3D skeletons, eliminating scale ambiguity at the cost of calibration dependence and sensitivity to occlusion. 
Despite sharing components like segmentation and tracking~\cite{sam3, bytetrack}, these two paradigms are rarely compared head-to-head on the same motion metrics.

\myparagraph{Evaluation without ground truth.}
Validating 3D reconstruction without ground truth requires indirect signals. Prior work uses anthropometric plausibility (e.g., bone-length consistency) to evaluate pose validity~\cite{wandt2019repnet}. Reprojection and cross-view self-consistency enable evaluation without 3D supervision~\cite{chen2019unsupervised, wandt2021canonpose}, a standard approach in egocentric capture~\cite{wang2021estimating}. We bundle these signals into a reference-free evaluation protocol tailored to urban pedestrian analysis.


\section{Dataset and Task Definition}


We evaluate both pipelines on StreetAware~\cite{streetaware}, a public multimodal dataset (RGB, audio, LiDAR) of urban intersections totaling nearly eight hours, each instrumented with four sensor pairs (one camera per corner) yielding partially overlapping multi-view coverage.
We restrict our analysis to the RGB modality and use three recording sessions from the dataset. Sessions~1 and~3 are longer, high-density recordings, whereas Session~2 is shorter and sparser.

Given this data, our task is to recover time-varying 3D body keypoints and trajectories for the pedestrians crossing the intersection. 
From these reconstructions, we derive the descriptors that matter most for pedestrian-centric analysis: anthropometric quantities such as stature and limb lengths, and kinematic quantities such as walking speed and direction.
Both the monocular and multi-view pipelines target these same outputs, enabling a direct, task-matched comparison.

Because such datasets rarely contain ground-truth 3D pose annotations or cross-camera pedestrian identity labels, evaluation is challenging.
It rules out conventional error-based evaluation. We therefore assess the reconstructed outputs using a reference-free protocol grounded in deployment-relevant criteria: pedestrian coverage, anthropometric plausibility relative to population statistics~\cite{fryar2016anthropometric}, walking-speed distributions, computational cost, and qualitative failure modes.


\section{Reconstruction Pipelines}

\figPipelines

We compare two pipelines producing the same outputs: 3D COCO-17 body keypoints and pedestrian trajectories. 
Both were configured toward the same goal: maximizing the fidelity of the recovered physical descriptors, such as stature, limb lengths, trajectories, and walking speed.
We tuned each to a comparable level of effort so that the comparison reflects the two strategies rather than uneven optimization. 
As summarized in \autoref{fig:pipelines}, Pipeline~A follows a monocular approach, while Pipeline~B follows a calibrated multi-view geometric strategy. 
The shared COCO-17 skeleton makes coverage, morphology, motion, and failure modes directly comparable.

\subsection{Pipeline A: Monocular Learned Reconstruction}
Pipeline~A estimates 3D body pose from a single fixed camera by integrating text-prompted instance segmentation, multi-object tracking, and learned 3D body estimation. Each frame is processed by SAM3~\cite{sam3} using the prompt ``\texttt{person}'', which produces instance masks, bounding boxes, and confidence scores (\autoref{fig:pipelines}, stage~2).

Detections with confidence below $\tau_\text{det}=0.40$ are discarded. Per-frame detections are associated over time with ByteTrack~\cite{bytetrack}, and masks are re-matched to tracks using maximum Intersection-over-Union to assign each pedestrian a persistent ID, corresponding to the third stage of Pipeline~A in \autoref{fig:pipelines}.

For each tracked instance, SAM3D Body~\cite{sam3dbody} receives the cropped RGB region and mask and regresses a parametric body with 127 anatomical landmarks and a weak-perspective camera translation $\mathbf{t}_{\mathrm{cam}}\in\mathbb{R}^3$. A fixed mapping projects these landmarks onto the COCO-17 skeleton, ensuring output comparability with Pipeline~B. This reconstruction process corresponds to the final stage of Pipeline~A in \autoref{fig:pipelines}.

Because the predicted translation is expressed in a camera-relative weak-perspective scale, a metric correction is applied using the camera intrinsic and extrinsic parameters.
The intrinsic matrix is rescaled to the inference resolution, depth is corrected using the calibrated focal length, and principal-point offsets are applied to the image-plane coordinates. 
The corrected camera-space point is then mapped to the global frame as
\begin{equation}
    \mathbf{p}_{\mathrm{world}} =
    \mathbf{R}_{c}^{\top}\mathbf{p}'_{\mathrm{cam}} + \mathbf{t}_{c}.
\end{equation}

This correction enables analysis of stature, limb lengths, and trajectories in metric units while preserving the monocular characteristics of the pose estimator.
The upper-torso center of mass serves as a representative point of the reconstructed pose for speed calculation.
Instantaneous speed is estimated from consecutive displacements of this representative point.
Stature is computed as the vertical span of the mapped COCO keypoints. 
Lastly, right arm and leg lengths are obtained by summing the corresponding shoulder–elbow–wrist and hip–knee–ankle segment lengths. 
To reduce depth jitter and enforce per-tracklet morphological consistency, stature and limb lengths are summarized by their median over each pedestrian trajectory.

\subsection{Pipeline B: Multi-View Geometric Reconstruction}
Pipeline~B reconstructs 3D pose from synchronized calibrated cameras through four stages: \textit{extrinsic calibration}, \textit{2D pose estimation with local tracking}, \textit{graph-based cross-camera identity matching}, and \textit{DLT triangulation}. Unlike Pipeline~A, it relies on projective geometry rather than learned depth priors, yielding metric 3D coordinates when calibration and cross-camera association are reliable.

The reconstruction setting is based on four camera pairs. In each pair, one camera serves as an anchor with a known global pose, while the pose of the target camera is estimated using Scale-Invariant Feature Transform (SIFT) correspondences filtered by Random Sample Consensus (RANSAC). The relative rotation and translation are determined from the essential matrix, and the baseline scale is established using static environmental reference points. 
This process yields projection matrices for all cameras:
\begin{equation}
    \mathbf{P}_{c} =
    \mathbf{K}_{c}
    \left[
    \mathbf{R}_{c} \mid \mathbf{t}_{c}
    \right].
\end{equation}

This calibration setup corresponds to the first stage of Pipeline~B in \autoref{fig:pipelines}.
Each camera stream is processed independently using YOLOv11-pose~\cite{yolo11}, with a relaxed detection threshold of $\tau_\text{conf} \geq 0.35$ to maintain coverage under occlusion and motion blur. ByteTrack links detections into local tracklets within each view, corresponding to the second stage of Pipeline~B in \autoref{fig:pipelines}.

To merge local tracklets into global identities, tracklet pairs with sufficient temporal overlap are compared using geometric validation. For each candidate pair, selected keypoints are triangulated over sampled concurrent frames. A match is accepted if the median reprojection error is below $\tau_{\mathrm{error}}=15$ pixels and the reconstructed head-to-ankle distance lies within $1.0$--$2.2$ m. The reprojection threshold was set empirically to balance recall under partial occlusion against the risk of false cross-camera associations. Accepted pairwise matches form an undirected graph, where connected components define global identities, with the constraint that no component contains two tracklets from the same camera, corresponding to the third stage of Pipeline~B in \autoref{fig:pipelines}.

For spatial reconstruction, a COCO keypoint is triangulated only if it is detected with a confidence of at least $0.30$ in two or more synchronized views. 
For each 2D observation $\mathbf{x}_{c}=(u_{c},v_{c})$, the DLT system is constructed as
\begin{equation}
(u_{c}\mathbf{P}_{c,3}-\mathbf{P}_{c,1})\mathbf{X}=0,
\qquad
(v_{c}\mathbf{P}_{c,3}-\mathbf{P}_{c,2})\mathbf{X}=0.
\end{equation}
This system is solved using Singular Value Decomposition. This triangulation process corresponds to the final stage of Pipeline~B in \autoref{fig:pipelines}.

Reconstructed joints located outside the 30-meter capture radius or outside the elevation range $-0.5 \leq z \leq 3.0$ meters are discarded. To eliminate severe tracking anomalies, trajectories with a median stature greater than $2.20$ meters are also excluded. Given the same keypoint structure, post-processing uses the same descriptors as Pipeline~A to calculate anthropometric measurements.


\section{Results and Analysis}
\label{sec:results}

Throughout this section, we report bootstrap 95\% confidence intervals for all median estimates (percentile method, $n_{\text{boot}}=10{,}000$ resamples). For stature—the only metric with an established population reference—we additionally evaluate distributional agreement against CDC adult reference data using a two-sample Kolmogorov--Smirnov (KS) test, which measures the maximum distance $D$ between cumulative distributions. To quantify the magnitude of deviation without non-normality assumptions, we report Cliff's $\delta$, a rank-based effect size ranging from $-1$ (complete downward shift) to $+1$ (complete upward shift).

\subsection{Pedestrian Coverage and Re-Identification}
Pipeline~A recovers more pedestrian tracklets because it operates from a single view and does not require simultaneous multi-camera visibility. On the reference camera alone, it produces 3,425, 909, and 5,278 tracklets for Sessions~1--3, respectively.

Pipeline~B initially generates many per-camera tracklets but retains only those that can be geometrically validated across views. Per-camera counts range from 760 to 1,520 in Session~1, from 182 to 353 in Session~2, and from 941 to 2,588 in Session~3, resulting in 1,883--4,461 tracklets per camera across all sessions. After cross-camera re-identification, which requires each trajectory to be matched in at least two cameras, the retained global identities drop to 372, 82, and 424 for Sessions~1--3, respectively, totaling 878 pedestrians.


This reduction reflects the cost of enforcing multi-view geometric validity rather than detection failure: Pipeline~B sacrifices coverage for reliable metric reconstruction, while Pipeline~A maximizes recall by accepting single-view trajectories—illustrating the central deployment trade-off between broad capture and validated 3D measurement.

\subsection{Anthropometric plausibility}
\label{subsec:height}
A central promise of learned 3D estimators is to recover body-scale descriptors from monocular video with minimal sensing infrastructure. \autoref{fig:morfologia} presents the reconstructed morphological metrics, comparing the right arm- and leg-length distributions from Pipeline~A (left) against the corresponding estimates from Pipeline~B (right).
As a biological baseline we model the adult population stature as an equal mixture of male and female normals parameterized from CDC anthropometric reference data~\cite{fryar2016anthropometric} (means $1.75$\,m and $1.61$\,m).

\figMorphology
\figStatureComparison

Pipeline~A yields a stable, physically bounded distribution (median $1.332$\,m, 95\% CI $[1.329, 1.334]$,  IQR $1.27$--$1.38$\,m) but systematically shifted downward relative to the reference, as shown in \autoref{fig:stature_comparison} (left).
This reflects the scale ambiguity intrinsic to monocular regression: lacking an absolute scale reference, the depth backbone relies on learned structural priors, and at large camera-to-subject distances ($z\approx 30$\,m in Session~1) with oblique pitch the weak-perspective projection compresses absolute stature.
As illustrated in \autoref{fig:stature_comparison} (right), Pipeline~B attains a comparable median ($1.435$\,m, 95\% CI $[1.414, 1.452]$) but, under its relaxed multi-view configuration, exhibits far larger variance (IQR $1.32$--$1.57$\,m) with implausible tails (down to $0.31$\,m and up to $2.20$\,m).
Although the tracklet association step enforces a head-to-ankle distance constraint ($1.0$\,m to $2.2$\,m) on a sample of frames during Re-ID matching, the subsequent multi-view triangulation of the full trajectory does not constrain stature frame-by-frame. 
Consequently, keypoint noise, mis-detections, and occlusions in individual frames propagate to the overall median stature of a few trajectories, leading to these degraded tails.
These extremes also stem from cross-camera keypoint mis-associations---e.g., matching the head of one individual to the ankles of another in dense frames---which propagate through triangulation as large geometric errors.
%

Pipeline~A is therefore useful for relative pose and tracking rather than absolute metric measurement, while Pipeline~B offers more plausible metric estimates for correctly associated tracklets but remains vulnerable to catastrophic outliers from cross-camera mismatches.

Bootstrap intervals and KS tests confirm both pipelines differ from the CDC reference ($p<0.001$)~\cite{fryar2016anthropometric}. However, Pipeline~B's distribution is notably closer to this baseline (median $1.435$\,m, 95\% CI $[1.414, 1.452]$; KS $D=0.65$, $\delta=-0.69$) than Pipeline~A's (median $1.332$\,m, 95\% CI $[1.329, 1.334]$; KS $D=0.97$, $\delta=-1.00$). Pipeline~A's extreme effect size ($\delta=-1.00$) confirms systematic height compression. For limb lengths, bootstrap intervals show high precision: Pipeline~B yields a right arm length of $0.532$\,m $[0.525, 0.537]$ and right leg of $0.796$\,m $[0.788, 0.801]$, whereas Pipeline~A yields $0.482$\,m $[0.480, 0.483]$ and $0.745$\,m $[0.744, 0.747]$, respectively.

%


In short, learned priors regularize scale at the cost of compression, while multi-view geometry recovers absolute scale but remains outlier-prone—motivating hybrid, skeletal-constrained triangulation.

\subsection{Motion dynamics}
\label{subsec:motion}
Extreme velocity values indicate residual tracking, triangulation, or
identity-association errors; velocity distributions are thus useful not only as mobility descriptors but also as diagnostic indicators of reconstruction quality. 
For each tracklet we adopt the median instantaneous speed of its upper-torso center of mass.
\autoref{fig:velocity_comparison} compares the two pipelines.
\figSpeedAnalysis

Pipeline~A has a median of $1.328$\,m/s with a pronounced right tail exceeding $30$\,m/s, reaching a maximum of $40$\,m/s, driven by ByteTrack identity switches that concatenate distinct individuals and by weak-perspective depth jitter at large distances.
Pipeline~B is more realistic and bounded (median $1.452$ \,m/s, 95\% CI $[1.448, 1.455]$; mean $2.645$\,m/s), with density tightly concentrated near the median and a markedly lighter tail than Pipeline~A; its residual tail arises from triangulation instability when pedestrians cross camera boundaries at poor intersection angles.
To assess the practical severity of these velocity outliers, we quantify the proportion of physically implausible estimates at two granularities using a threshold of $12.4$\,m/s---the fastest sprinting speed ever recorded for a human---with Wilson-score 95\% confidence intervals. At the frame level, $1.36$\% [95\% CI: $1.27$--$1.45$] of Pipeline~A instantaneous-speed estimates and $4.05$\% [95\% CI: $3.93$--$4.17$] of Pipeline~B estimates exceed this threshold. Because we report tracklet-level median speed throughout this paper, the more relevant figure is at the trajectory level: only $1.52$\% [95\% CI: $1.19$--$1.93$] of Pipeline~A tracklets and $0.11$\% [95\% CI: $0.02$--$0.64$] of Pipeline~B tracklets ($1$ of $878$) exceed it, indicating that per-tracklet median aggregation already filters out most frame-level noise for Pipeline~B.
Leveraging explicit cross-camera constraints, Pipeline~B mitigates the depth jitter that affects monocular regression, but the persistent tail in both pipelines highlights the shared need for robust temporal filtering and plausibility constraints.

\subsection{Processing time}
\label{subsec:processing_time}
All experiments ran on a single NVIDIA Quadro RTX~6000 GPU (24\,GB). 
Pipeline~A, on one camera stream, processes each frame in about $4.3$\,s on average---roughly $48$\,h for a $40{,}000$-frame sequence---split between SAM~3 segmentation and SAM~3D body estimation.
Pipeline~B is lightweight: it processes about $980{,}000$ frames across the three sessions in $539.4$\,min (${\approx}9$\,h), i.e.,\ ${\approx}33$\,ms/frame (${\approx}30$\,fps) for YOLOv11-pose, its main bottleneck (${\approx}8$\,h total).
Cross-camera association and global Re-ID solve a multi-view graph (e.g.\ $5{,}409$ nodes and $13{,}442$ edges in Session~1; $8{,}980$ nodes and $25{,}625$ edges in Session~3), identifying $878$ unique pedestrians in ${\approx}45$\,min.
Overall, Pipeline B achieves a $130\times$ speedup per frame per camera over Pipeline A. Although Pipeline B operates over eight simultaneous camera streams, its total wall-clock time remains far lower than running Pipeline A on even a single stream at the same scale, making it the only viable option for real-time or large-corpus deployments.

This gap should not be interpreted as an inherent property of the monocular or multi-view paradigms. Instead, it reflects the computational cost of the specific architectures compared: Pipeline A relies on heavyweight promptable segmentation-and-mesh-recovery models (SAM~3 and SAM~3D Body), whereas Pipeline B is built on a lightweight 2D pose estimator (YOLOv11-pose). A lighter monocular estimator would likely narrow this gap substantially, so the reported speedup should be read as evidence about the tested implementations rather than as a general claim that multi-view reconstruction is inherently faster than monocular reconstruction.


\section{Practical Recommendations}

To guide practitioners in selecting the appropriate reconstruction framework in urban intersection deployments where 3D ground truth is unavailable, we distill the trade-offs observed in \autoref{sec:results} into four core operational criteria.

\myparagraph{Infrastructure Constraints.}
When the deployment site lacks synchronized, calibrated cameras (or when the existing network consists of legacy, non-overlapping views) Pipeline~A is the only viable option. 
Pipeline~B's cross-camera Re-ID stage requires geometric validation across at least two synchronized views; in its absence, the graph-based association stage cannot be executed and reliable multi-view metric reconstruction cannot be performed.
This flexibility makes it the natural choice for opportunistic or resource-constrained deployments.

\myparagraph{Real-Time Edge Analytics.}
When throughput and latency are primary constraints, Pipeline~B is preferred in the tested configuration. As shown in \autoref{subsec:processing_time}, Pipeline~B achieved an inference rate of approximately 33~ms per frame. In contrast, Pipeline~A's processing time of 4.3~s per frame makes it approximately 130$\times$ slower than Pipeline~B. This gap renders Pipeline~A impractical for real-time or large-corpus deployments without substantial additional hardware resources.
As discussed in \autoref{subsec:processing_time}, this speed difference reflects the relative computational cost of the specific models evaluated (SAM~3 and SAM~3D Body vs.\ YOLOv11-pose) rather than an inherent advantage of multi-view geometry over monocular estimation, and could narrow considerably with a lighter monocular estimator.

\myparagraph{Traffic Safety and Kinematics.}
Applications that demand absolute metric trajectories—such as time-to-collision (TTC) estimation, conflict zone analysis, or speed-limit compliance auditing—require Pipeline~B. For anthropometric quantities, Pipeline~B's central stature distribution aligns more closely with CDC reference data, as shown in \autoref{subsec:height}, and, for correctly associated tracklets, provides calibrated metric estimates suitable for safety-critical inference. Pipeline~A's more pronounced height compression and scale ambiguity make its absolute metric outputs unreliable for such use cases without an explicit scale correction step. 
As quantified in \autoref{subsec:motion}, only $1.52$\% [95\% CI: $1.19$--$1.93$] of Pipeline~A tracklets and $0.11$\% [95\% CI: $0.02$--$0.64$] of Pipeline~B tracklets have a median speed exceeding the $12.4$\,m/s plausibility threshold; nonetheless, we recommend that any safety-critical deployment apply an explicit speed-plausibility filter at this threshold prior to computing kinematic safety metrics such as TTC, since even a small fraction of contaminated trajectories can produce spurious near-collision alarms.

\myparagraph{Crowd Flow and Relative Motion Analysis.}
When the goal is maximizing pedestrian recall for crowd-flow analysis, occupancy estimation, or relative pose studies, Pipeline~A is preferred.
Because it requires only a single view, it avoids the substantial data attrition introduced by Pipeline~B's cross-camera Re-ID filter: across all sessions, Pipeline~B retained only 878 unique global identities after Re-ID out of thousands of per-camera tracklets, a reduction that reflects the strict geometric validity constraint rather than true pedestrian absence. 
Pipeline~A recovers many more single-view tracklets, albeit with relative rather than fully reliable absolute body measurements. Its learned structural priors also yield a stable, physically bounded morphological distribution that is suitable for relative comparisons across crowd segments, even if absolute scale cannot be trusted.


\section{Conclusion}

This paper presented a reference-free comparison of monocular learned and multi-view geometric strategies for deriving pedestrian motion descriptors from urban intersection video. Our results highlight a clear trade-off: the monocular approach (Pipeline A) provides broad coverage and stable structural bounds, but suffers from scale ambiguity and height compression. Conversely, the geometric approach (Pipeline B) yields fast, metric, and realistic mobility estimates, achieving a $130\times$ speedup, but remains vulnerable to calibration dependence, data attrition, and outliers from cross-camera misassociation. While limited by the lack of absolute 3D ground-truth error metrics, this study provides practical guidance for configuring sensing infrastructure according to the target pedestrian-dynamics analysis task. 
This study is further limited by its reliance on a single dataset comprising three sessions from one urban intersection; although our empirical observations show consistent trends across these sessions, validating these findings across different camera configurations, intersection geometries, and pedestrian densities remains important future work. Similarly, a fine-grained stratification of reconstruction quality by pedestrian--camera distance, occlusion severity, and instantaneous scene density -- beyond the session-level and threshold-based analyses reported here -- would further strengthen the evaluation and is left for future work. Future work will explore hybrid frameworks that inject learned structural priors into multi-view triangulation, combining the metric accuracy of geometric methods with the physical plausibility of learned models.
\section*{Acknowledgments}
This work was partially supported by the Brazilian National Council for Scientific and Technological Development (CNPq, grant \#313454/2026-4 and \#132348/2025-0), the Carlos Chagas Filho Foundation for Research Support of the State of Rio de Janeiro (FAPERJ, grants E-26/210.585/2025, E-26/204.005/2025, and E-26/204.004/2025), the São Paulo Research Foundation (FAPESP, grant \#2021/07012-0), and the School of Applied Mathematics at Fundação Getulio Vargas.
\bibliographystyle{IEEEtran}
\bibliography{references}
%
%
\end{document}
/ 6
Pipeline A maximizes recall by accepting single-view trajec-
tories—illustrating the central deployment trade-off between
broad capture and validated 3D measurement.
B. Anthropometric plausibility
A central promise of learned 3D estimators is to recover
body-scale descriptors from monocular video with minimal
sensing infrastructure. Fig. 2 presents the reconstructed mor-
phological metrics, comparing the right arm- and leg-length
distributions from Pipeline A (left) against the corresponding
estimates from Pipeline B (right). As a biological baseline we
model the adult population stature as an equal mixture of male
and female normals parameterized from CDC anthropometric
reference data [20] (means 1.75 m and 1.61 m).
Pipeline A yields a stable, physically bounded distribution
(median 1.332 m, 95
but systematically shifted downward relative to the reference,
as shown in Fig. 3 (left). This reflects the scale ambiguity
intrinsic to monocular regression: lacking an absolute scale
reference, the depth backbone relies on learned structural
priors, and at large camera-to-subject distances (z ≈ 30 m
in Session 1) with oblique pitch the weak-perspective pro-
jection compresses absolute stature. As illustrated in Fig. 3
(right), Pipeline B attains a comparable median (1.435 m,
95
configuration, exhibits far larger variance (IQR 1.32–1.57 m)
with implausible tails (down to 0.31 m and up to 2.20 m).
Although the tracklet association step enforces a head-to-
ankle distance constraint (1.0 m to 2.2 m) on a sample of
frames during Re-ID matching, the subsequent multi-view
triangulation of the full trajectory does not constrain stature
frame-by-frame. Consequently, keypoint noise, mis-detections,
and occlusions in individual frames propagate to the overall
median stature of a few trajectories, leading to these degraded
tails. These extremes also stem from cross-camera keypoint
mis-associations—e.g., matching the head of one individual
to the ankles of another in dense frames—which propagate
through triangulation as large geometric errors.
Pipeline A is therefore useful for relative pose and tracking
rather than absolute metric measurement, while Pipeline B
offers more plausible metric estimates for correctly associated
tracklets but remains vulnerable to catastrophic outliers from
cross-camera mismatches.
Bootstrap intervals and KS tests confirm both pipelines
differ from the CDC reference (p < 0.001) [20]. However,
Pipeline B’s distribution is notably closer to this baseline
(median 1.435 m, 95
δ = −0.69) than Pipeline A’s (median 1.332 m, 95
[1.329, 1.334]; KS D = 0.97, δ = −1.00). Pipeline A’s
extreme effect size (δ = −1.00) confirms systematic height
compression. For limb lengths, bootstrap intervals show
high precision: Pipeline B yields a right arm length of
0.532 m [0.525, 0.537] and right leg of 0.796 m [0.788, 0.801],
whereas Pipeline A yields 0.482 m [0.480, 0.483] and 0.745 m
[0.744, 0.747], respectively.
In short, learned priors regularize scale at the cost of
compression, while multi-view geometry recovers absolute
scale but remains outlier-prone—motivating hybrid, skeletal-
constrained triangulation.
C. Motion dynamics
Extreme velocity values indicate residual tracking, trian-
gulation, or identity-association errors; velocity distributions
are thus useful not only as mobility descriptors but also
as diagnostic indicators of reconstruction quality. For each
tracklet we adopt the median instantaneous speed of its upper-
torso center of mass. Fig. 4 compares the two pipelines.
Pipeline A has a median of 1.328 m/s with a pronounced
right tail exceeding 30 m/s, reaching a maximum of 40 m/s,
driven by ByteTrack identity switches that concatenate distinct
individuals and by weak-perspective depth jitter at large dis-
tances. Pipeline B is more realistic and bounded (median 1.452
m/s, 95
tightly concentrated near the median and a markedly lighter
tail than Pipeline A; its residual tail arises from triangulation
instability when pedestrians cross camera boundaries at poor
intersection angles. To assess the practical severity of these
velocity outliers, we quantify the proportion of physically
implausible estimates at two granularities using a threshold
of 12.4 m/s—the fastest sprinting speed ever recorded for
a human—with Wilson-score 95
the frame level, 1.36
instantaneous-speed estimates and 4.05
of Pipeline B estimates exceed this threshold. Because we
report tracklet-level median speed throughout this paper, the
more relevant figure is at the trajectory level: only 1.52%
[95
CI: 0.02–0.64] of Pipeline B tracklets (1 of 878) exceed it,
indicating that per-tracklet median aggregation already filters
out most frame-level noise for Pipeline B. Leveraging explicit
cross-camera constraints, Pipeline B mitigates the depth jitter
that affects monocular regression, but the persistent tail in
both pipelines highlights the shared need for robust temporal
filtering and plausibility constraints.
D. Processing time
All experiments ran on a single NVIDIA Quadro RTX 6000
GPU (24 GB). Pipeline A, on one camera stream, processes
each frame in about 4.3 s on average—roughly 48 h for a
40,000-frame sequence—split between SAM 3 segmentation
and SAM 3D body estimation. Pipeline B is lightweight: it
processes about 980,000 frames across the three sessions in
539.4 min (≈9 h), i.e., ≈33 ms/frame (≈30 fps) for YOLOv11-
pose, its main bottleneck (≈8 h total). Cross-camera associa-
tion and global Re-ID solve a multi-view graph (e.g. 5,409
nodes and 13,442 edges in Session 1; 8,980 nodes and 25,625
edges in Session 3), identifying 878 unique pedestrians in
≈45 min. Overall, Pipeline B achieves a 130× speedup per
frame per camera over Pipeline A. Although Pipeline B
operates over eight simultaneous camera streams, its total wall-
clock time remains far lower than running Pipeline A on even
0.0 0.2 0.4 0.6 0.8 1.0 1.2 1.4 1.6
Meters
0.0
2.5
5.0
7.5
10.0
Density
Right Arm Length (m)
Estimated
Estimated KDE
median = 0.539 m
(95
0.0 0.2 0.4 0.6 0.8 1.0 1.2 1.4 1.6
Meters
0.0
2.5
5.0
7.5
10.0
Density
Right Leg Length (m)
Estimated
Estimated KDE
median = 0.798 m
(95
Integrated Morphological Distributions (Arm & Leg)0.0 0.2 0.4 0.6 0.8 1.0 1.2 1.4 1.6
Meters
0.0
2.5
5.0
7.5
10.0
Density
Right Arm Length (m)
Estimated
Estimated KDE
median = 0.482 m
(95
0.0 0.2 0.4 0.6 0.8 1.0 1.2 1.4 1.6
Meters
0.0
2.5
5.0
7.5
10.0
Density
Right Leg Length (m)
Estimated
Estimated KDE
median = 0.745 m
(95
Integrated Morphological Distributions (Arm & Leg)0.0 0.2 0.4 0.6 0.8 1.0 1.2 1.4 1.6
Meters
0.0
2.5
5.0
7.5
10.0
Density
Right Arm Length (m)
Estimated
Estimated KDE
median = 0.539 m
(95
0.0 0.2 0.4 0.6 0.8 1.0 1.2 1.4 1.6
Meters
0.0
2.5
5.0
7.5
10.0
Density
Right Leg Length (m)
Estimated
Estimated KDE
median = 0.798 m
(95
Integrated Morphological Distributions (Arm & Leg)0.0 0.2 0.4 0.6 0.8 1.0 1.2 1.4 1.6
Meters
0.0
2.5
5.0
7.5
10.0
Density
Right Arm Length (m)
Estimated
Estimated KDE
median = 0.482 m
(95
0.0 0.2 0.4 0.6 0.8 1.0 1.2 1.4 1.6
Meters
0.0
2.5
5.0
7.5
10.0
Density
Right Leg Length (m)
Estimated
Estimated KDE
median = 0.745 m
(95
Integrated Morphological Distributions (Arm & Leg)Right Arm Length (m)
Pipeline A Pipeline B
Right Leg Length (m)
Pipeline A Pipeline BFigure 2. Comparison of reconstructed morphological metrics (right leg length, top; right arm length, bottom) between Pipeline A (left) and Pipeline B (right).
The distributions are estimated using a per-tracklet global median assignment, with dashed lines indicating the median values.0.0 0.5 1.0 1.5 2.0
Meters
0.0
1.5
3.0
4.5
6.0
Density
Total Stature (m)
Estimated
Estimated KDE
median = 1.332 m
(95
Ref. Pop.
Distribution: Total Stature (m)0.0 0.5 1.0 1.5 2.0
Meters
0.0
1.5
3.0
4.5
6.0
Density
Total Stature (m)
Estimated
Estimated KDE
median = 1.435 m
(95
Ref. Pop.
Distribution: Total Stature (m)Stature Distribution (m)
Figure 3. Comparison of estimated total stature distributions across all three
sessions between Pipeline A (monocular learned, left) and Pipeline B (multi-
view geometric, right). The dashed curve representing the reference population
(Ref. Pop.) models the adult population stature as an equal mixture of male and
female normals parameterized from CDC data. Dashed vertical lines indicate
the median of the estimated distributions.0 5 10 15 20 25 30 35 40
Average speed (m/s)
0.0
0.1
0.2
0.3
0.4
0.5
Density
Estimated
Estimated KDE
median = 1.452 m/s
(95
World Record = 12.4 m/s (4.05
Integrated Pedestrian Speed0 5 10 15 20 25 30 35 40
Average speed (m/s)
0.0
0.1
0.2
0.3
0.4
0.5
Density
Estimated
Estimated KDE
median = 1.328 m/s
(95
World Record = 12.4 m/s (1.36
Integrated Pedestrian SpeedSpeed Distribution (m/s)
Figure 4. Comparison of estimated pedestrian speed distributions between
Pipeline A (left) and Pipeline B (right). Histograms show the estimated density
with overlaid kernel density estimates (KDE); dashed lines indicate the median
speed values (1.328 m/s for Pipeline A and 1.452 m/s for Pipeline B).
a single stream at the same scale, making it the only viable
option for real-time or large-corpus deployments.
This gap should not be interpreted as an inherent property
of the monocular or multi-view paradigms. Instead, it reflects
the computational cost of the specific architectures compared:
Pipeline A relies on heavyweight promptable segmentation-
and-mesh-recovery models (SAM 3 and SAM 3D Body),
whereas Pipeline B is built on a lightweight 2D pose estimator
(YOLOv11-pose). A lighter monocular estimator would likely
narrow this gap substantially, so the reported speedup should
be read as evidence about the tested implementations rather
than as a general claim that multi-view reconstruction is
inherently faster than monocular reconstruction.
VI. PRACTICAL RECOMMENDATIONS
To guide practitioners in selecting the appropriate recon-
struction framework in urban intersection deployments where
3D ground truth is unavailable, we distill the trade-offs ob-
served in Section V into four core operational criteria.
Infrastructure Constraints. When the deployment site lacks
synchronized, calibrated cameras (or when the existing net-
work consists of legacy, non-overlapping views) Pipeline A is
the only viable option. Pipeline B’s cross-camera Re-ID stage
requires geometric validation across at least two synchronized
views; in its absence, the graph-based association stage cannot
be executed and reliable multi-view metric reconstruction
cannot be performed. This flexibility makes it the natural
choice for opportunistic or resource-constrained deployments.
Real-Time Edge Analytics. When throughput and latency
are primary constraints, Pipeline B is preferred in the tested
configuration. As shown in Section V-D, Pipeline B achieved
an inference rate of approximately 33 ms per frame. In
contrast, Pipeline A’s processing time of 4.3 s per frame
makes it approximately 130× slower than Pipeline B. This
gap renders Pipeline A impractical for real-time or large-
corpus deployments without substantial additional hardware
resources. As discussed in Section V-D, this speed difference
reflects the relative computational cost of the specific models
evaluated (SAM 3 and SAM 3D Body vs. YOLOv11-pose)
rather than an inherent advantage of multi-view geometry over
monocular estimation, and could narrow considerably with a
lighter monocular estimator.
Traffic Safety and Kinematics. Applications that demand
absolute metric trajectories—such as time-to-collision (TTC)
estimation, conflict zone analysis, or speed-limit compliance
auditing—require Pipeline B. For anthropometric quantities,
Pipeline B’s central stature distribution aligns more closely
with CDC reference data, as shown in Section V-B, and,
for correctly associated tracklets, provides calibrated metric
estimates suitable for safety-critical inference. Pipeline A’s
more pronounced height compression and scale ambiguity
make its absolute metric outputs unreliable for such use cases
without an explicit scale correction step. As quantified in
Section V-C, only 1.52
tracklets and 0.11
metries, and pedestrian densities remains important future work. Similarly, a fine-grained stratification of reconstruction quality by pedestrian--camera distance, occlusion severity, and instantaneous scene density -- beyond the session-level and threshold-based analyses reported here -- would further strengthen the evaluation and is left for future work. Future work will explore hybrid frameworks that inject learned structural priors into multi-view triangulation, combining the metric accuracy of geometric methods with the physical plausibility of learned models.

\section*{Acknowledgments}

This work was partially supported by the Brazilian National Council for Scientific and Technological Development (CNPq, grant \#313454/2026-4 and \#132348/2025-0), the Carlos Chagas Filho Foundation for Research Support of the State of Rio de Janeiro (FAPERJ, grants E-26/210.585/2025, E-26/204.005/2025, and E-26/[Kauan]), the São Paulo Research Foundation (FAPESP, grant \#2021/07012-0), and the School of Applied Mathematics at Fundação Getulio Vargas.



\bibliographystyle{IEEEtran}
\bibliography{references}
%
%


\end{document}